\documentclass[letterpaper, 10 pt, conference]{ieeeconf}  % Comment this line out if you need a4paper

\IEEEoverridecommandlockouts                              

\usepackage{graphics} % for pdf, bitmapped graphics files
\usepackage{multirow} % for cmd 'multirow', 'multicolumn'
\usepackage{amsmath,amsfonts}
\usepackage{algorithmicx}
\usepackage{algpseudocode}
\usepackage{array}
\usepackage[caption=false,font=footnotesize,labelfont=rm,textfont=rm]{subfig}
\usepackage{textcomp}
\usepackage{stfloats}
\usepackage{url}
\usepackage{verbatim}
\usepackage{graphicx}
\usepackage[noadjust]{cite}
\usepackage{amsmath} % assumes amsmath package installed
\usepackage{amssymb}  % assumes amsmath package installed
\usepackage[ruled,linesnumbered]{algorithm2e}
\usepackage{subfig}
\usepackage{bm}
\title{Infra-Swarm: Robust Vision-Based Multi-Robot Swarming\\ via Near-Infrared Spectral Vision
}

\author{Haoyu Chen$^{*1}$, Qijin Li$^{*2}$, Wanyu Xiang$^{3}$, Xiuxiu Lin$^{1}$, Zian Ning$^{\dagger 1}$, and Shiyu Zhao$^{\dagger 1}$% <-this % stops a space
\thanks{This research work was supported by National Natural Science Foundation of China (Grant No. 62473320).}%
\thanks{*These authors contributed equally to this work. $^{\dagger}$Joint corresponding author.}%
\thanks{$^{1}$H. Chen, X. Lin, Z. Ning, and S. Zhao are with the WINDY Lab in the School of Engineering at Westlake University, Hangzhou, China. S. Zhao is also with the Research Center for Industries of the Future, Westlake University, Hangzhou, China. 
\{chenhaoyu, linxiuxiu, ningzian, zhaoshiyu\}@westlake.edu.cn}%
\thanks{$^{2}$Q. Li is with The Hong Kong University of Science and Technology (Guangzhou), Guangzhou, China}%
\thanks{$^{3}$W. Xiang is with Zhejiang University, Hangzhou, China.}%
}

\begin{document}

\newcommand{\I}{\mathbf{I}}

\renewcommand{\O}{\Omega}
\newcommand{\Obar}{\bar{\Omega}}
\newcommand{\sat}{\mathrm{sat}}
\renewcommand{\d}{\mathrm{d}}

\newcommand{\blue}{\textcolor{blue}}
\newcommand{\red}{\textcolor{red}}

\newcommand{\A}{\mathbf{A}}
\renewcommand{\a}{\mathbf{a}}
\newcommand{\B}{\mathbf{B}}
\newcommand{\e}{\mathbf{e}}
\newcommand{\F}{\mathbf{F}}

\newcommand{\g}{\mathbf{g}}
\newcommand{\G}{\mathcal{G}}
\renewcommand{\H}{\mathbf{H}}
\newcommand{\h}{\mathbf{h}}
\newcommand{\E}{\mathbb{E}}
\newcommand{\J}{\mathcal{J}}

\newcommand{\M}{\mathbf{M}}
\newcommand{\N}{\mathcal{N}}

\renewcommand{\P}{\mathbf{P}}
\newcommand{\p}{\mathbf{p}}
\newcommand{\R}{\mathbf{R}}
\renewcommand{\S}{\mathbf{S}}
\newcommand{\s}{\mathbf{s}}

\newcommand{\U}{\mathbf{U}}
\newcommand{\smallu}{\mathbf{u}}
\newcommand{\V}{\mathcal{V}}
\newcommand{\vel}{\mathbf{v}}
\newcommand{\Vel}{\mathbf{V}}
\newcommand{\q}{\mathbf{q}}
\newcommand{\f}{\mathbf{f}}
\newcommand{\bk}{\mathbf{K}}

\newcommand{\w}{\mathbf{w}}
\newcommand{\W}{\mathbf{W}}
\newcommand{\x}{\mathbf{x}}
\newcommand{\y}{\mathbf{y}}
\newcommand{\z}{\mathbf{z}}

\newcommand{\zero}{\mathbf{0}}

\newcommand{\Eta}{\boldsymbol{\eta}}
\newcommand{\bomega}{\boldsymbol{\omega}}
\newcommand{\btau}{\boldsymbol{\tau}}

\graphicspath{{figures/}}

\maketitle
\thispagestyle{empty}
\pagestyle{empty}

%%%%%%%%%%%%%%%%%%%%%%%%%%%%%%%%%%%%%%%%%%%%%%%%%%%%%%%%%%%%%%%%%%%%%%%%%%%%%%%%
\begin{abstract}
Distributed swarms typically rely on either active wireless communication or passive vision, and they are frequently hindered by bandwidth constraints or environmental sensitivity. This paper proposes \emph{Infra-Swarm}, a robust vision-based swarm. Each robot is equipped with a near-infrared light source and four ordinary gray-scale cameras. The Infra-Swarm system directly measures the centimeter-level 3D position of neighbors based on the position (bearing) and intensity (strength) of optical flares in the captured images. By utilizing 940 nm narrow-band filters to physically reject 99.2\% of ambient light interference, the perception front-end achieves hardware-level robustness against illumination variations. Furthermore, its minimal computational overhead provides a resilient foundation for the massive scalability of robotic collectives on resource-constrained hardware. 
\end{abstract}
%%%%%%%%%%%%%%%%%%%%%%%%%%%%%%%%%%%%%%%%%%%%%%%%%%%%%%%%%%%%%%%%%%%%%%%%%%%%%%%%
\section{Introduction}
Multi-robot swarms hold significant potential for applications such as environmental exploration~\cite{Burgard2000} and cooperative transportation~\cite{Sun2023,zhu2025multi}. At the heart of these distributed systems lies neighbor interaction through consistent information exchange, typically achieved via wireless communication or visual perception. 

Wireless communication remains the predominant collaborative approach, offering high-bandwidth real-time sharing. However, its active nature fundamentally constrains large-scale scalability due to time-division and frequency-division multiplexing~\cite{gielis2022critical}. As node density increases, time-division induces exponential growth in latency, while frequency-division leads to spectral resource exhaustion, both of which severely limit the realization of massive swarms.

Alternatively, conventional vision-based approaches provide passive sensing without bandwidth bottlenecks. Natural biological swarms such as starling murmurations and sardine shoals utilize vision to effortlessly achieve colossal scales of tens of thousands of individuals~\cite{King2012Murmurations,Pavlov2000FishSchooling}. These biological systems have inspired a variety of passive vision-based solutions, with researchers developing specialized frameworks for underwater~\cite{Berlinger2021FishSwarm}, terrestrial~\cite{Mezey2025Land}, and aerial robots~\cite{Chung2018Aerial}. Theoretically, vision-enabled distributed systems bypass channel competition through independent perception, thereby inherently supporting massive scalability.

However, current visual perception systems face three critical challenges:
1) Limited environmental adaptability: Traditional RGB-based detectors are highly susceptible to atmospheric and lighting variables. In degraded outdoor environments, these algorithms encounter catastrophic failures, with the average F-measure plunging to 0.2881 under poor illumination~\cite{Roy2021}. This lack of robustness creates a critical bottleneck for large-scale deployments, where reliable neighbor perception is essential but frequently obstructed by diurnal cycles and shadows.
2) High computational complexity: Deep learning-based detectors such as YOLO improve accuracy, and foundation models enhance scene understanding, yet they demand substantial onboard computing resources. Even with optimized hardware accelerators, real-time inference consumes approximately 18 W per node~\cite{Nguyen2019}. This creates prohibitive cost barriers for large-scale deployments, as a 100-robot swarm would require 1.8 kW of computational power alone.
3) Limited information dimensions: While existing vision-only systems can estimate neighbors' kinematic states (position/velocity)~\cite{dai2017scannet}, they lack semantic information exchange capabilities. This represents a critical disadvantage compared to wireless communication, which enables transmission of task instructions and environmental data beyond basic motion states, severely constraining collaborative task flexibility.

\begin{figure}[!t]
    \centering
    \includegraphics[width=1\linewidth]{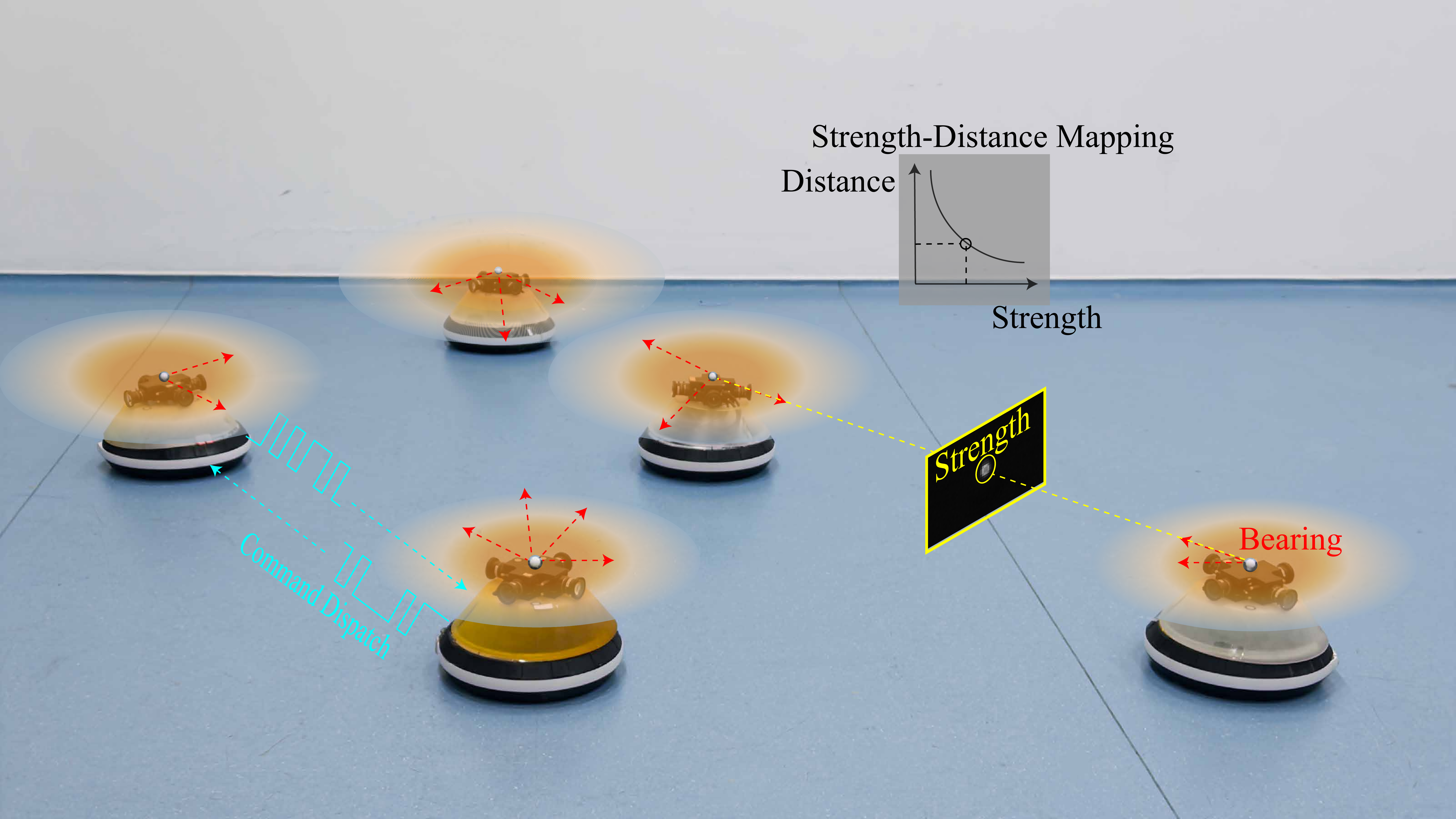}
    \caption{The Infra-Swarm paradigm enables \textbf{robust} relative localization and semantic exchange by decoding near-infrared flares. In this framework, bearing and radiometric strength serve as primary observations for 3D state estimation, while temporal modulation of the light source facilitates integrated optical communication.}
    \label{fig:sketch map}
\end{figure}

To address these challenges, we present Infra-Swarm, a distributed robotic swarm system based on Near-Infrared Spectral Vision. As illustrated in Fig~\ref{fig:hardware}, each robot integrates a 940~nm near-infrared light source (NILS) and four industrial-grade grayscale cameras equipped with narrowband optical filters featuring a bandwidth of less than 10nm.

Our dual-mode system combines photometric ranging and optical language communication to achieve:
1) High robustness perception: Narrowband filtering eliminates 99.2\% ambient light interference, maintaining $>$99.8\% detection success across 300-800~lux illumination conditions. Unlike active RGB markers whose signal-to-noise ratio is severely degraded by broad-spectrum ambient light, this narrowband isolation creates a pristine optical channel, effectively achieving the illumination robustness of foundation models but with strictly $\mathcal{O}(1)$ computational efficiency.
2) Low-cost system: Leveraging the inverse-square law of light intensity, our direct ranging method achieves centimeter-level accuracy with a Mean Absolute Error of 2.8 cm. The system enables single-frame processing in 10 ms, which is 30$\times$ faster than conventional vision algorithms when running on Raspberry Pi-level hardware.
3) Multimodal communication: Through LED intensity modulation at a frequency of 10Hz, we encode semantic information such as motion commands and task allocation into optical pulse sequences. This approach effectively breaks the information-dimensionality barrier inherent in vision-only systems.

The primary contribution of this work lies in the sensing-communication integrated hardware stack. To objectively validate this perception front-end, we deliberately employ a standard pseudo-linear Kalman filter. The proposed system incorporates three key technical innovations:

First, we designed an omnidirectional radiation source that integrates diffuser encapsulation with constant-current drivers, which reduces light intensity variation to below 10\% across $\pm85^\circ$ angular range. This performance significantly outperforms conventional LED solutions, exhibiting over 40\% intensity fluctuation, thereby ensuring millimeter-level consistency in full-space distance measurement. 
% source 不均匀，加了diffusor，定性定量 性能/对比

Second, we developed a multispectral vision architecture employing four 940~nm filtered monochrome cameras to achieve a $360^\circ \times 90^\circ$ field of view with a 10~m effective tracking range. By physically rejecting 99.2\% of ambient light interference, it establishes hardware-level robustness against diverse illumination conditions. Furthermore, the system ensures inherent scalability through $\mathcal{O}(1)$ perception: a single image-processing pass detects all visible neighbors simultaneously, completely decoupling the computational overhead from the swarm density.

Third, we developed an optical language communication protocol that implements a UART-based asynchronous decoding algorithm. This protocol demonstrates exceptional error resilience, sustaining less than 0.1\% bit error rate when processing 30-frame-per-second video streams, while simultaneously supporting concurrent transmission of 16 distinct commands to dynamically moving targets.

Experimental results demonstrate successful autonomous formation control in five-robot swarms with an inter-robot spacing error of less than 5~cm. Compared to existing solutions, the infra-swarm system achieves an 80\% reduction in hardware costs compare to normal vision-based swarm robots, which are brought to less than 500\$ per node. This efficiency paves the way for industrial inspection, disaster response, and other large-scale swarm applications.

\section{Related Works}
Existing methods can be categorized into three groups based on the mode of information sharing: wireless communication-based, vision-based, and marker-based approaches.
% 第三类

\subsection{Wireless Communication-Based Swarms}
Wireless communication remains the most widely adopted paradigm for neighbor interaction~\cite{Vasarhelyi2018}. Systems utilizing Ultra-Wideband or dedicated radio protocols offer high-precision state estimation and real-time information exchange for small to medium-scale groups~\cite{Sun2023}. However, as the swarm expands, these active communication-based methods encounter fundamental scalability constraints. The primary limitation arises from the underlying multiplexing mechanisms: Time-Division Multiple Access leads to an exponential increase in communication latency as more nodes compete for broadcasting slots, while Frequency-Division Multiple Access faces severe spectral resource exhaustion in high-density deployments. These bottlenecks inevitably degrade the synchronization and stability of the swarm, necessitating a more scalable, independent perception-based alternative that bypasses channel competition.

\subsection{Vision-Based Swarms}
Inspired by biological collectives~\cite{King2012Murmurations}, researchers have explored passive vision to bypass the channel competition inherent in wireless communication, aiming for massive scalability through independent perception~\cite{Berlinger2021FishSwarm,Mezey2025Land, Chung2018Aerial}. Recent advancements have further enhanced positioning precision via binocular stereo vision~\cite{Cheng2025Binocular} and enabled agile flight through differentiable physics-based learning~\cite{Zhang2025Learning}. 

However, standard passive vision—regardless of whether it employs multi-camera geometric reconstruction~\cite{Cheng2025Binocular} or end-to-end deep learning~\cite{Zhang2025Learning, Bochkovskiy2020}—faces critical deployment barriers. The fundamental reliance on ambient light distribution renders detection reliability highly susceptible to volatile or poor illumination, which often leads to catastrophic perception failures. Furthermore, the substantial matching overhead of binocular systems and the inference costs of learning-based controllers necessitate high-performance onboard computing, thereby precluding the use of resource-constrained hardware in large-scale collectives. Beyond these physical and computational constraints, passive vision typically lacks a mechanism for semantic exchange, restricting agents to basic kinematic tracking~\cite{li2025collective}. These multi-dimensional limitations motivate the development of active marker systems that reconcile visual scalability with signal-level robustness.

\subsection{Active Markers and Near-Infrared Spectral Vision}
To enhance perception in adverse conditions, specialized tracking algorithms like ADTrack~\cite{Li2021ADTrack} leverage dual-filter learning to maintain stability in dark environments. However, such passive methods remain computationally intensive and fundamentally constrained by ambient light distribution. To achieve higher robustness, active marker systems such as UVDAR~\cite{Walter2019} and its communication-integrated successor UVDAR-COM~\cite{horyna2022} utilize ultraviolet (UV) LEDs to physically isolate robot signals from complex backgrounds. While these UV-based systems provide exceptional resilience against ambient interference, they primarily rely on geometric detection or frequency-based blinking to distinguish agents. Consequently, they face significant limitations in resolving overlapping visual projections; when multiple light sources coincide in the image plane, the resulting signal coalescence leads to ambiguous localization and increased processing demands.

In contrast, Infra-Swarm introduces a radiometric ranging approach. By applying the inverse-square law to 940~nm near-infrared light sources, we directly map pixel intensities to physical distances. This RSS-inspired methodology not only reduces computational costs relative to ADTrack but also uniquely leverages intensity variations to decouple overlapping targets. Even when visual projections coincide, the distinct radiometric signatures allow for reliable depth separation, ensuring consistent localization in dense, high-speed swarms.

\begin{figure*}[!t]
    \centering
    \includegraphics[width=\linewidth]{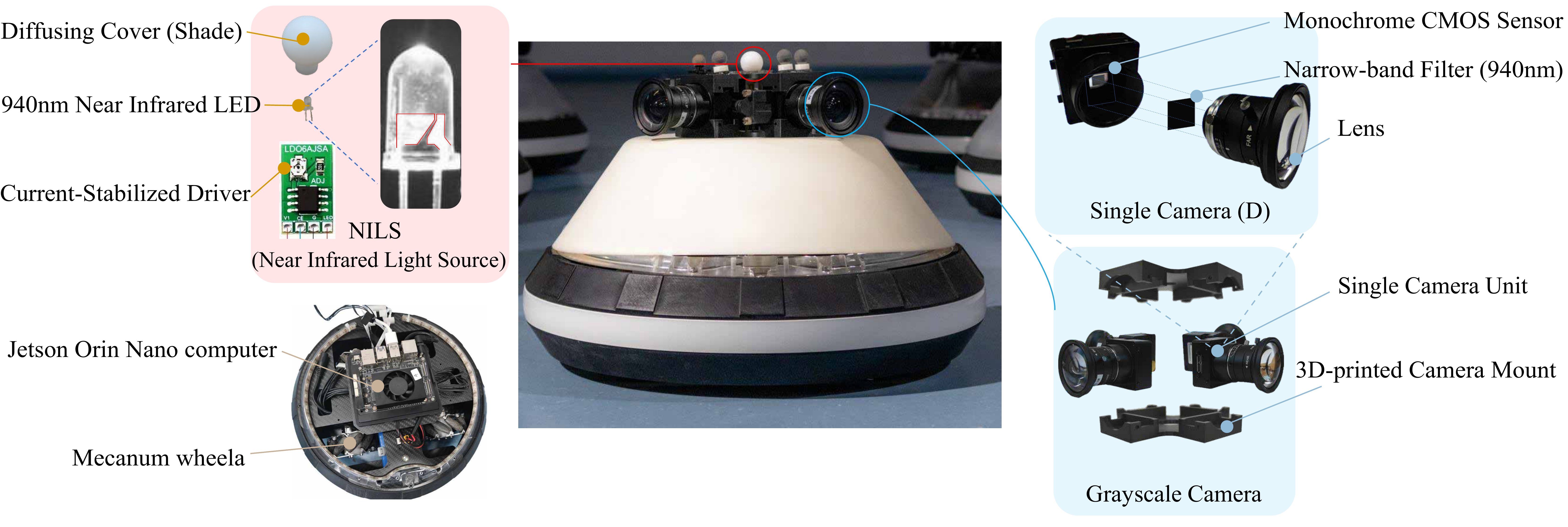}
    \caption{The hardware platform integrates a $360^{\circ}$ infrared sensing suite onto an omnidirectional mobile base.}
    \label{fig:hardware}
\end{figure*}
\section{Hardware System Design} 
\label{s:System}
Figure~\ref{fig:hardware} depicts the Infra-Swarm hardware, featuring an omnidirectional mobile base, a NILS with a diffusing shade, and a four-camera monochrome array with narrow-band filtering. 

\subsection{Base Platform and Hardware Transition}
\label{ss:Hardware_Overview}
%引用图2
The system is developed on the Omnibot platform~\cite{Ma2024}, as illustrated in the center of Fig.~\ref{fig:hardware}. The platform features an omnidirectional chassis driven by four Mecanum wheels and is equipped with an onboard Jetson Orin Nano computer. It is worth noting that our framework is highly light-weight; the entire perception and localization pipeline is implemented solely on the CPU of the Orin Nano without GPU acceleration, demonstrating its compatibility with resource-constrained hardware such as Raspberry Pi-class processors. The internal configuration of the sensing suite is detailed in the bottom-left inset of Fig.~\ref{fig:hardware}. To mitigate environmental lighting interference, we replaced the original RGB cameras with this specialized infrared-based setup, allowing the robot to isolate navigation signals from the visible spectrum and bypass the non-linearities inherent in consumer-grade vision hardware.
%解释一句只用了CPU
\subsection{Near-Infrared Light Source}
\label{ss:NILS}
The NILS is designed for a high signal-to-noise ratio and emission consistency. We selected the 940~nm near-infrared wavelength to minimize interference from solar radiation and avoid spectral overlap with motion capture systems like VICON (operating at 850nm). This wavelength also aligns with the peak quantum efficiency (QE) of standard CMOS sensors. To ensure spatial uniformity, each LED is encased in a custom diffusing cover, reducing intensity variation from over 40\% to less than 10\% across all viewing angles. For temporal stability, a current-stabilized driver circuit with a DC-DC buck converter is implemented. This design decouples light emission from battery voltage fluctuations, ensuring a constant radiant flux essential for precise distance modeling.

\subsection{Infrared-Optimized Monochrome Camera Array}
\label{ss:Camera}
The perception system utilizes four Mindvision industrial monochrome cameras, configured to provide a 360-degree field of view. These cameras utilize a CMOS sensor with a QE exceeding 15\% at the 940~nm wavelength, enabling the detection of 0.2W NILS emitters at distances up to 10m. To manage the wide dynamic range of signal intensities, an adaptive exposure control mechanism is implemented via direct hardware control to prevent pixel saturation from proximal robots or signal loss at long ranges. Crucially, the system bypasses internal ISP pipelines to access raw grayscale data directly. This preserves the photometric fidelity of the captured flares, which is essential for accurate radiometric ranging.

\begin{figure*}[!t]
    \centering
    \includegraphics[width=\linewidth]{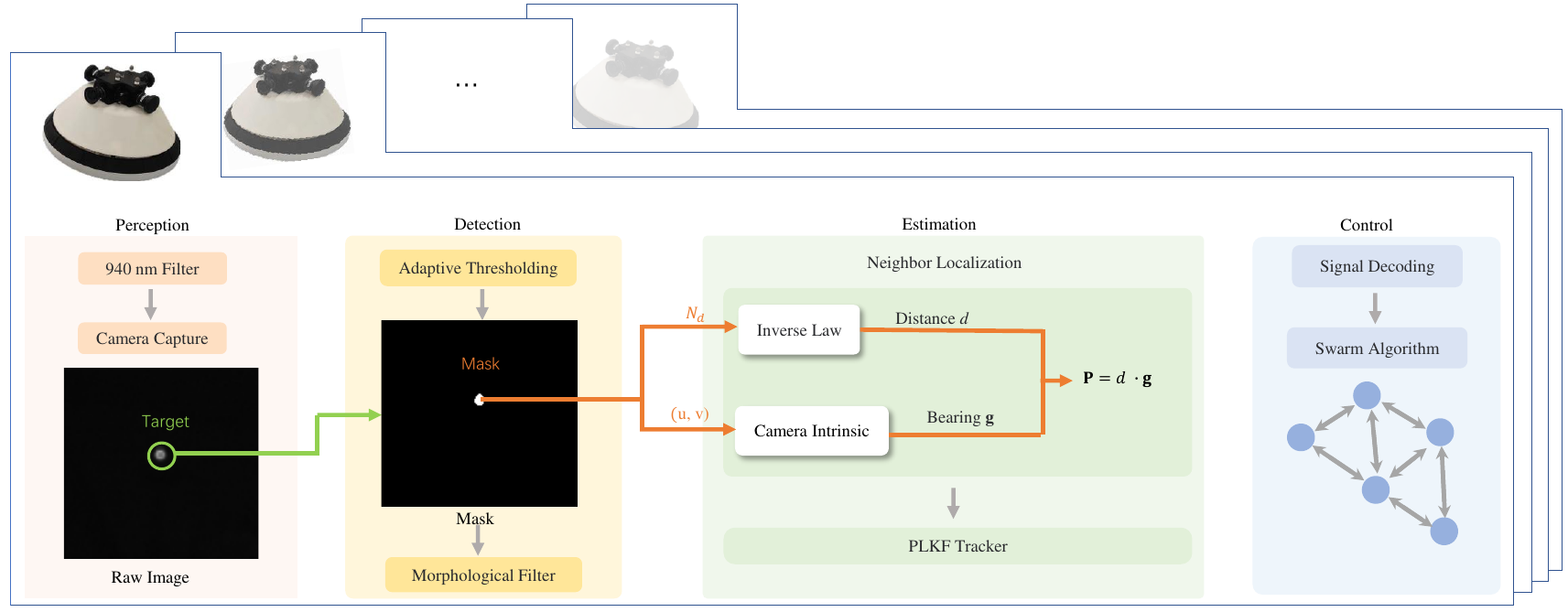}
    \caption{The overview of the Infra-Swarm algorithmic framework.}
    \label{fig:pipeline}
\end{figure*}

\section{Methodology}
\label{s:method}
Figure~\ref{fig:pipeline} illustrates the computational framework of Infra-Swarm, which integrates infrared feature extraction with a pseudo-linear measurement model for robust 3D neighbor tracking.
\subsection{Grayscale-based Ranging}
\label{ss:grayscale_ranging_algo}
The core of the proposed localization framework is the radiometric ranging model, which directly maps the raw pixel intensity to the physical distance $d$. Based on the inverse square law of light, the distance is derived as
\begin{align}
\label{eq:distance}
d = \sqrt{\frac{K \cdot t}{N_{\textup{d}} - \bar{N}}},
\end{align}
where $N_{\textup{d}}$ represents the average grayscale value of the detected flare, $\bar{N}$ denotes the local background noise, $t$ is the camera exposure time, and $K$ is a lumped system constant determined via calibration.

The derivation of \eqref{eq:distance} relies on the \textit{isotropic point source assumption}, which posits that the emitter acts as a dimensionless point radiating energy uniformly in all directions. Since the operational distance $d$ is significantly larger than the physical dimension of the NILS emitter (i.e., $d \gg r_{\text{source}}$), the light source is modeled as a point emitter. Combined with the diffusing cover described in Section~\ref{ss:NILS}, the irradiance $E$ incident on the sensor follows the inverse square law given by
\begin{align}
\label{eq:irradiance}
E = \frac{\Phi}{4\pi d^2} + E_{\text{env}},
\end{align}
where $\Phi$ is the radiant flux and $E_{\text{env}}$ is the ambient radiation. Furthermore, by bypassing the non-linear ISP, the camera maintains a strictly linear photometric response. The recorded pixel value $N_{\textup{d}}$ is thus directly proportional to the accumulated photon energy, expressed as
\begin{align}
\label{eq:camera_response}
N_{\textup{d}} = K_{\text{c}} \frac{t S}{f_{\text{s}}^2} E,
\end{align}
where $K_{\text{c}}$ is the sensor response coefficient, $S$ is the effective pixel area, and $f_{\text{s}}$ is the focal length. Substituting \eqref{eq:irradiance} into \eqref{eq:camera_response} yields the definition of the system constant 
\begin{align*}
K = \frac{K_{\text{c}} S \Phi}{4\pi f_{\text{s}}^2},
\end{align*}
which completes the mathematical derivation.

To obtain the necessary inputs $N_{\textup{d}}$ and $\bar{N}$ for \eqref{eq:distance}, we employ a streamlined image processing pipeline designed for raw grayscale data. The process begins with adaptive thresholding to isolate high-intensity signals from environmental illumination. The segmentation threshold is dynamically set to twice the global mean grayscale value of the current frame, ensuring robustness against varying ambient light levels. From the resulting binary mask, connected components are extracted and subjected to geometric filtering to reject noise. Only candidates satisfying specific constraints—specifically, a pixel area $A_k > 4$ and an aspect ratio between $0.8$ and $1.2$—are validated as NILS markers. For each confirmed target, the average signal intensity $N_{\textup{d}}$ is computed from the pixels within the flare contour, while the local background noise $\bar{N}$ is sampled from the immediate vicinity. These extracted values serve as the real-time inputs for the distance calculation described in Section~\ref{ss:grayscale_ranging_algo}.
\subsection{Directional Bearing Estimation}
To supplement the scalar range $d$, the unit bearing vector $\mathbf{g}$ in the robot's body coordinate frame is derived through back-projection of the detected flare's centroid. For a target center at pixel coordinates $(u, v)$, we first compute the normalized direction vector in the image plane using the inverse of the camera intrinsic matrix $\mathbf{K}_{\text{cam}}$:
\begin{equation}
\label{eq:pixel_to_image}
\mathbf{p}_{\text{img}} = \frac{\mathbf{K}_{\text{cam}}^{-1} [u, v, 1]^\top}{\|\mathbf{K}_{\text{cam}}^{-1} [u, v, 1]^\top\|},
\end{equation}
where $\mathbf{K}_{\text{cam}}$ is pre-determined via calibration. The vector $\mathbf{p}_{\text{img}}$ is then transformed into the 3D camera coordinate system through a fixed rotation matrix $\mathbf{R}_{\text{i2c}}$ that accounts for the sensor's internal orientation:
\begin{equation}
\mathbf{p}_{\text{cam}} = \mathbf{R}_{\text{i2c}} \mathbf{p}_{\text{img}}.
\end{equation}

To achieve a unified perception space for the panoramic multi-camera setup, the local camera-frame coordinates are mapped to the robot's body frame (car coordinates) using the extrinsic transformation matrix $\mathbf{T}_{c}^{b}$ corresponding to the specific camera ID:
\begin{equation}
\label{eq:cam_to_car}
\mathbf{p}_{\text{body}} = \mathbf{T}_{c}^{b} \begin{bmatrix} \mathbf{p}_{\text{cam}} \\ 1 \end{bmatrix}.
\end{equation}
Finally, the unit bearing vector $\hat{\mathbf{g}}$ used in the 3D tracking framework is obtained by normalizing the translational components of $\mathbf{p}_{\text{body}}$. This process ensures that detections from all four cameras are fused into a consistent ego-centric coordinate system, providing the necessary angular inputs for the subsequent PLKF-based state estimation.
\subsection{3D Multi-target Tracking and Overlap Resolution}
% 伪线性
Building upon the independent detection capabilities of each camera, we propose a multi-target tracking framework based on a 3D Pseudo-linear Kalman Filter (PLKF)~\cite{li2022three}. This approach mitigates two critical challenges inherent in dense swarm environments: identity (ID) switches arising from overlapping 2D projections of distinct 3D trajectories, and intermittent signal occlusions caused by the high-frequency modulation of the LED emitters. By elevating the tracking state from the 2D image plane to a unified 3D physical coordinate system~\cite{zhang2026observability}, the proposed framework effectively resolves spatial ambiguities.

To estimate the kinematic state of each target, we employ a linear constant-velocity model. The state vector $\mathbf{x}_t$ at time $t$ is defined as the relative position $\delta_{\mathbf{p}}$ and velocity $\delta_{\mathbf{v}}$ between the target and the observer:
\begin{equation}
    \mathbf{x}_t = \begin{bmatrix}\delta_{\mathbf{p}} \\ \delta_{\mathbf{v}} \end{bmatrix}, \quad
    \text{with } \delta_{\mathbf{p}} = \mathbf{p}_T - \mathbf{p}_o, \enspace \delta_{\mathbf{v}} = \mathbf{v}_T - \mathbf{v}_o.
\end{equation}
Since external control inputs are unknown, the state evolution follows a linear transition model:
\begin{equation}
    \mathbf{x}_{t} = F \mathbf{x}_{t-1} + B\mathbf{q}_{t-1},
\end{equation}
where the transition matrix $F$ and noise gain $B$ incorporate the sampling interval $\delta t$:
\begin{equation}
    F = \begin{bmatrix} \mathbf{I}_3 & \delta t \cdot \mathbf{I}_3 \\ \mathbf{0}_3 & \mathbf{I}_3 \end{bmatrix}, \quad
    B = \begin{bmatrix} \frac{1}{2}\delta t^2 \cdot \mathbf{I}_3 \\ \delta t \cdot \mathbf{I}_3 \end{bmatrix}.
\end{equation}

While the state dynamics are linear, the raw measurements consist of a nonlinear bearing vector $\hat{\mathbf{g}}$ and a scalar range $\hat{r}$, corrupted by directional noise $\boldsymbol{\mu} \sim \mathcal{N}(0, \sigma_\mu^2 \mathbf{I}_3)$ and range noise $w \sim \mathcal{N}(0, \sigma_w^2)$~\cite{ning2024bearing}. To avoid the derivative-based linearization and the associated Jacobian complexity of an Extended Kalman Filter (EKF), we introduce a projection matrix $P_{\hat{\mathbf{g}}} = \mathbf{I}_3 - \hat{\mathbf{g}} \hat{\mathbf{g}}^\top$. This allows us to formulate a pseudo-linearized measurement equation $\mathbf{z} = H \mathbf{x} + \boldsymbol{\nu}$, where:
\begin{equation}
    H = \begin{bmatrix}
    P_{\hat{\mathbf{g}}} & \mathbf{0}_{3\times3} \\
    \mathbf{I}_3 & \mathbf{0}_{3\times3}
    \end{bmatrix}, \quad
    \mathbf{z} = \begin{bmatrix}
    \mathbf{0}_{3} \\
    \hat{r} \hat{\mathbf{g}}
    \end{bmatrix}.
\end{equation}
The equivalent measurement noise covariance $R$ is derived specifically to account for the projection effects:
\begin{equation}
    R = \begin{bmatrix}
    \hat{r}^2 \sigma_\mu^2 P_{\hat{\mathbf{g}}} & \mathbf{0} \\
    \mathbf{0} & \sigma_w^2 \hat{\mathbf{g}} \hat{\mathbf{g}}^\top
    \end{bmatrix}.
\end{equation}
This formulation allows the use of standard linear Kalman update rules while rigorously handling the measurement geometry.

With the filter established, robust data association is performed following the observation-centric SORT\cite{Cao_2023_CVPR} pipeline, matching new observations $Z$ with existing trajectories $\hat{X}$ via the Hungarian Algorithm. We construct a composite cost matrix $\mathbf{C}$, where the cost between the $i$-th detection and $j$-th track combines 2D spatial overlap ($C_{\mathrm{IoU}}$), motion direction consistency ($C_{\theta}$), and a critical depth disparity term ($C_d$):
\begin{equation}
    C_{ij} = C_{\mathrm{IoU}} + C_{\theta} + w_d \left(1 - \exp\left(-| d^{\mathrm{det}}_i - d^{\mathrm{trk}}_j |\right)\right).
\end{equation}
The depth term $C_d$ ensures that targets overlapping in the image plane but separated in depth are correctly distinguished. Finally, to ensure continuity during intermittent signal loss, the tracker implements a robust lifecycle management strategy. This involves relying on prediction without correction when targets are temporarily lost, generating virtual trajectories via interpolation upon re-association to fill missing frames, and terminating tracks that remain unmatched for consecutive frames to prevent drift.

\section{Experimental Results}
\label{s:experiment}
Figure~\ref{fig_exp} illustrates our experimental evaluation, which progresses from two fundamental validation tests to two swarm coordination tasks.
\subsection{Exp1: Calibration and Ranging Accuracy}
Our theoretical model (Eq.~\eqref{eq:distance} and Eq.~\eqref{eq:camera_response}) shows that distance \(d\) can be found from image data if the system parameter \(k\) is known, as the term \(m = (N_\textup{d}-\bar{N})/t\) is directly measured. This subsection describes how we experimentally determine the constant \(k\).

\begin{figure*}[!t]
    \centering
    \subfloat[Exp1: Radiometric model verification via inverse square law fitting.]{
        \includegraphics[width=0.4\linewidth]{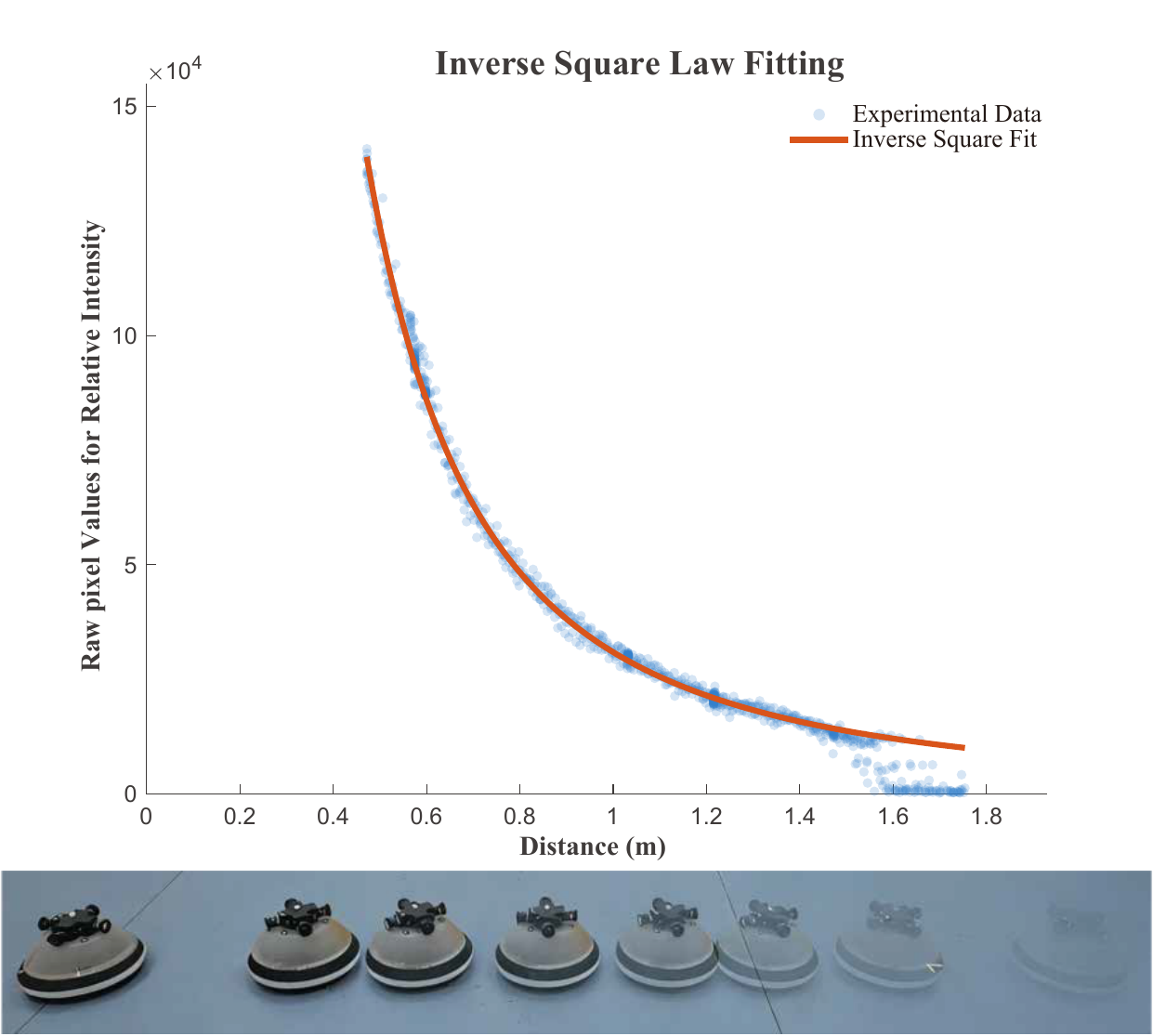} 
        \label{fig:calibration_overlay}
    }
    \subfloat[Exp2: Signal homogenization analysis with and without NILS diffusing shade.]{
        \includegraphics[width=0.55\linewidth]{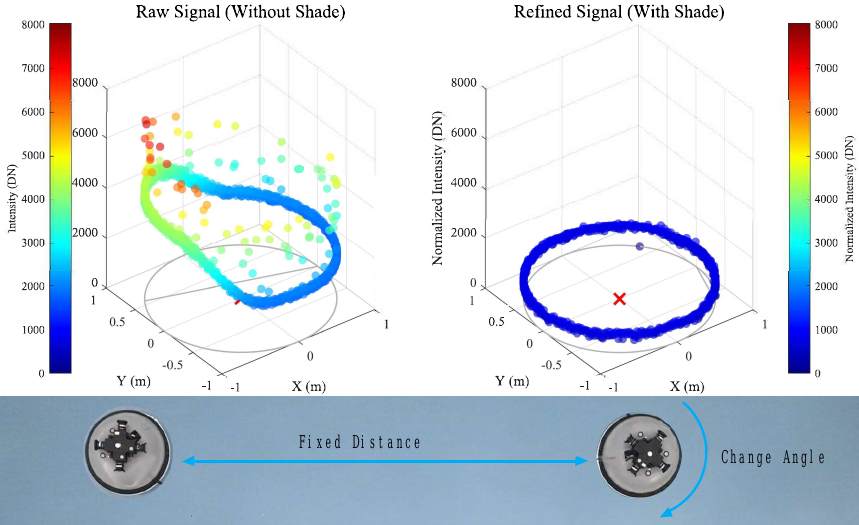}
        \label{fig:shade_comparison}
    }\\
    % \subfloat[Exp3: Snapshot of the physical five-robot swarm experiment.]{
    %     \includegraphics[width=0.37\linewidth]{Figures/swarm5.pdf}
    %     \label{fig:swarm}
    % }
    \begin{minipage}[b]{0.38\linewidth} % 这个宽度容纳 3.1 和 3.2
        \centering
        \subfloat[Exp 3.1:Swarm coordination in Scenario A.]{
            \includegraphics[width=0.42\linewidth]{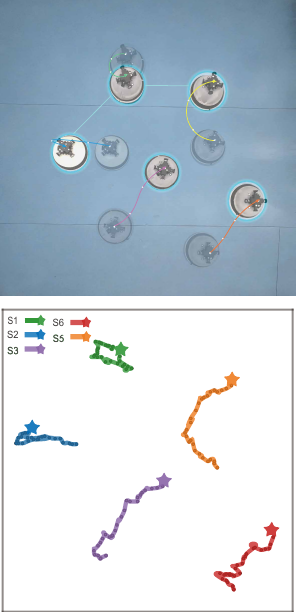}
            \label{fig:exp3_1}
        }\hspace{0.1em}
        \subfloat[Exp 3.2:Swarm coordination in Scenario B.]{
            \includegraphics[width=0.42\linewidth]{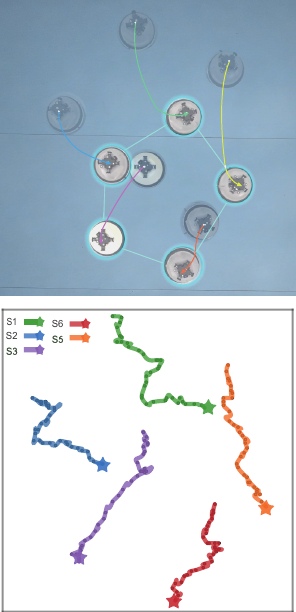}
            \label{fig:exp3_2}
        }
        % 如果你需要在这个小区域下方写总的 "Exp3"，可以取消下面这行的注释
         % \centerline{\footnotesize (c) Exp3: Five-robot swarm experiment.}
         
    \end{minipage}
    \hfill
    \subfloat[Exp4: Multi-stage formation reconfiguration for dynamic swarm coordination via infra-light communication]{
        \includegraphics[width=0.6\linewidth]{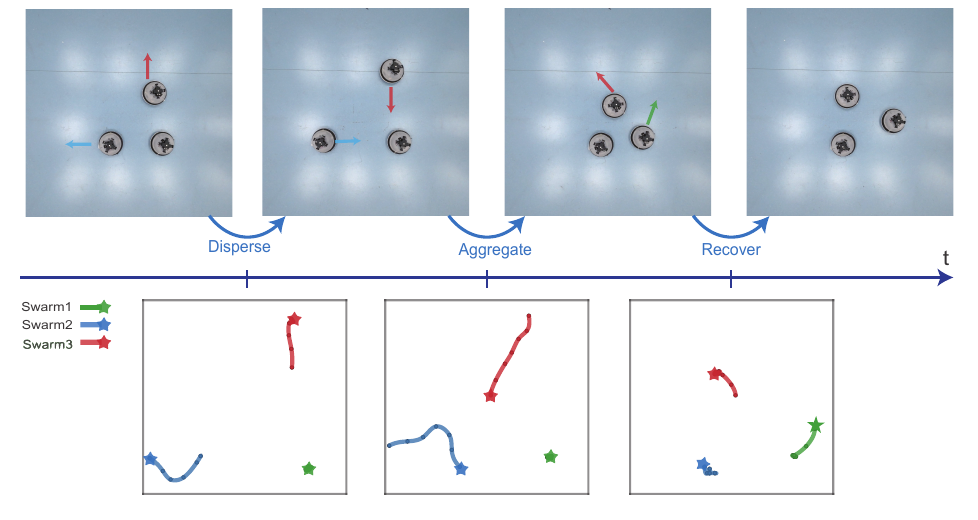}
        \label{fig:commu}
    }
    \caption{Experimental validation of the radiometric perception model and decentralized swarm coordination performance.}
    \label{fig_exp}
\end{figure*}

Figure~\ref{fig:calibration_overlay} illustrates the calibration setup, which used two robots from our swarm. One robot served as a mobile IR light source, and the other used its onboard camera to capture images of the source. A VICON motion capture system tracked both robots to provide precise ground-truth distances (\(d_{\text{VICON}}\)) between them. In the experiment, we moved the light-source robot to several known distances from the camera robot. At each distance, we captured multiple images. The camera's exposure time \(t\) was kept at a fixed, predetermined value throughout the entire calibration experiment.

We define the measured signal strength as $m = (N_{\textup{d}} - \bar{N})/t$. According to our radiometric model $d = \sqrt{k/m}$, a linear relationship exists such that $m = k \cdot d^{-2}$. By performing a linear regression of the empirical signal $m_i$ against the inverse square of the ground-truth distance $1/d_{\text{VICON},i}^2$, the system parameter $k$ is identified as the resulting slope. This calibration facilitates the direct mapping from raw pixel values to metric distance during autonomous operation. Figure~\ref{fig:calibration_overlay} illustrates this fitting concept.

% Trajectories of the robot swarm executing formation reconfigurations over three time intervals: 5--10~s (Disperse), 18--24~s (Aggregate), and 24--30~s (Recover). Swarm1 acts as the leader broadcasting infra-light commands. The star symbols ($\star$) denote the final positions at the end of each respective interval.

\subsection{Exp2: Point Source Verification}
\label{subsec:point_source_verification}
To validate the emission characteristics of our infrared (IR) light sources, specifically their conformity to the point source assumption utilized in Eq.~\eqref{eq:irradiance}, we conducted an angular uniformity test. The objective was to confirm that our custom IR emitters, equipped with diffusers, provide a sufficiently uniform intensity distribution for reliable operation. This experiment utilized the VICON motion capture system for precise angular measurement and a pair of our robotic platforms.

The experimental setup involved one robot serving as a stationary IR light source and a second robot, the observation robot, positioned at a fixed distance of 1~meter. The observation robot employed its onboard camera, with predetermined and fixed parameters, to capture images of the light source. The light-source robot was then precisely rotated about its vertical axis. During this rotation, the VICON system recorded the exact orientation of the emitter. Simultaneously, the observation robot captured images and performed real-time analysis of these images to determine the apparent intensity of the IR source at each recorded angle. This process allowed us to map the light source's intensity profile as a function of its rotation.

The results in Fig.~\ref{fig:shade_comparison} demonstrate that the raw LED exhibits a highly anisotropic emission pattern, with intensity fluctuations exceeding 40\% due to the inherently irregular internal geometry of the LED die and its packaging. Such non-uniformity would introduce catastrophic errors in a radiometric ranging model. In contrast, the refined signal with the diffusing shade presents a homogenized, near-Lambertian radiation profile. The shade effectively mitigates hardware-level spikes and dips, reducing the intensity variation to within $\pm$10\% across the operational angular range. This verification confirms that our NILS setup successfully transforms a directional emitter into a predictable point source, satisfying the requirements for the $1/d^2$ intensity fall-off model and enabling consistent 360\(^{\circ}\) perception.

\subsection{Exp3: Swarm Coordination Without External Tracking}
\label{ss:swarm_test}
% TODO: 修改，看看有没有需要补充的，可能得把一些放到method部分

To validate the integrated performance of the Infra-Swarm system, we conducted leader-follower flocking experiments in a 6$\times$8~m arena under approximately 500~lux ambient lighting. The swarm comprised one manually controlled leader and four autonomous followers. Each follower executes a distributed Reynolds-based flocking law to maintain formation based solely on onboard perception:
\begin{align}
\label{eq:swarm_control}
\mathbf{v}_i = \sum_{j \in \mathcal{N}_i} (\mathbf{v}_{\text{sep},ij} + \mathbf{v}_{\text{alg},ij} + \mathbf{v}_{\text{coh},ij}),
\end{align}

where \(\mathbf{v}_i\) and \(\mathbf{p}_i\) denote the velocity and position of robot \(i\), and \(\mathcal{N}_i\) represents its neighbors.

Furthermore, while line-of-sight occlusions are physically inevitable, they are inherently benign to our decentralized control topology. A distant robot is only occluded by a closer one; since the flocking laws prioritize nearest neighbors, continuously tracking the proximal occluding robot ensures macroscopic swarm safety.

The Robot Operating System (ROS) is utilized to implement the decentralized flocking logic, which aggregates three Reynolds-based \cite{olfati2006flocking}~behaviors through local perception. \textit{Separation} maintains a distance of $0.8$~m ($w=1.0$), \textit{alignment} matches velocity within $1.3$~m ($w=0.8$), and \textit{cohesion} ensures centering within $3.0$~m ($w=3.0$). The resultant velocity is limited to $0.1$~m/s to maintain system stability.

Figs.~\ref{fig:exp3_1} and~\ref{fig:exp3_2} showcase trajectories from a 120-second experimental trial. Even without external localization, the followers maintained a stable formation with a 1~m average distance. Although our system performs full 3D tracking, results are projected onto a 2D plane for visual clarity, as the ground robots operated at a uniform altitude. The 3D tracking framework, operating at 100~FPS, proved highly robust against the spatial ambiguities and signal occlusions inherent in dense formations. These results confirm that our radiometric-based pipeline delivers sufficiently precise state estimates for autonomous, decentralized swarming in unpredictable settings.

\subsection{Exp4: Infra-Light Communication System}
To enable robust decentralized coordination in radio frequency-restricted environments, we integrated an infra-light communication module that shares hardware with the visual positioning system. This dual-use design ensures scalability and electromagnetic interference resistance.

The transmission protocol is designed around a customized UART-based framing structure\cite{Wang2022} to ensure reliable data encoding under dynamic conditions. The transmitter employs a Jetson Edge computer to encode command data into discrete frames, where each frame comprises a start bit of 0, a 7-bit payload, an even parity bit, and two stop bits set to 1. The payload is capable of encoding either motion vectors or standard ASCII characters, depending on the operational context. To mitigate transmission errors during dynamic maneuvers, five identical frames are aggregated into a single transmission packet. This binary sequence is modulated by a microcontroller to drive infrared LEDs at a blinking frequency of $f=10$~Hz.The LED remains constantly lit when idle to maintain visual tracking continuity.

On the receiving end, an asynchronous decoding pipeline is implemented to recover signals without requiring hardware clock synchronization. The receiver utilizes four narrowband grayscale cameras capturing signals at $F=30$~Hz, a configuration that satisfies the sampling theorem with an oversampling ratio of $F=3f$.

This process involves three sequential stages, starting with sequence reconstruction. Due to asynchronous clocks, the number of samples per bit may vary. We apply a clustering algorithm to group consecutive identical samples. By utilizing the theoretical repetition factor ($F/f \approx 3$) and tolerance parameters, abnormal clusters are split to reconstruct a discrete binary sequence that aligns one-to-one with the transmitted bits.

Following reconstruction, frame synchronization and validation are performed to ensure data integrity. A sliding window algorithm is implemented to locate valid data frames within the reconstructed sequence. Upon detecting a signal transition (start bit $t_0$), the algorithm predicts the theoretical stop bit position $t_k$. A search window $[t_k-\delta, t_k+\delta+5]$ is then established to identify the stop bit pattern $\{1 \enspace 1\}$. This dynamic window effectively compensates for cumulative clock drift and sampling jitter.

Finally, data integrity is verified via even parity checks and frame separators $\{1 \enspace 1 \enspace 0\}$. If a check fails, the decoding window resets to the next valid start bit, preventing the system from processing noise as valid data.

To evaluate the practical utility of the infra-light communication system for swarm coordination, a multi-stage formation reconfiguration experiment was conducted with Swarm1 as the leader and Swarm2-3 as followers. As illustrated in Fig.~\ref{fig:commu}, the followers sequentially executed three distinct commands transmitted via the infra-light communication: Disperse, Aggregate, and  Recover. In the first phase, followers performed outward maneuvers to increase the average inter-robot distance from the initial 1.0~m to 1.2~m. In the second phase, the swarm adjusted its motion vectors to converge toward a local center, reducing the inter-robot spacing to 0.6~m. Finally, the robots readjusted their positions to restore the original 1.0~m spatial configuration. These results demonstrate that the proposed communication framework provides a robust link for real-time coordination, effectively enabling complex swarm maneuvers in infrastructure-free environments.

\section{Conclusion}
This paper demonstrates Infra-Swarm, a robust decentralized swarm system realized through integrated infra-visual sensing and localization. The primary advantage of this system is its exceptional stability: it maintains reliable performance under complex backgrounds, nocturnal conditions, and signal interference, while remaining resilient to partial occlusions—all while operating with low computational demand on CPU-class hardware. Furthermore, the framework enables seamless semantic exchange through temporal light modulation. Real-world experiments validate the collective stability of the swarm in dynamic scenarios. Future research will explore the scaling of this paradigm to ultra-large-scale collectives in unstructured environments.

However, we acknowledge that the current system evaluation establishes a baseline within a strictly defined operational envelope: a 5-robot collective operating under controlled indoor lighting, minimal environmental clutter, and restricted spatial ranges. Assessing the true limits of this architecture requires pushing beyond these boundaries. Scaling to massive collectives will inevitably introduce optical coalescence and temporal modulation collisions. Similarly, deploying in unstructured real-world environments will test the system's resilience against multi-path interference from reflective surfaces, dynamic range saturation under intense outdoor solar radiation, and severe diffuse reflection caused by atmospheric conditions such as fog or mist, which could introduce substantial estimation errors into the radiometric ranging model. Future work will focus on integrating advanced filtering algorithms and adaptive communication protocols into the broader robotics technology stack to systematically address these extreme deployment challenges.

\bibliography{myOwnPub} % if no referece is cited before, there will be an error
\bibliographystyle{ieeetr}

\end{document}